# Industrial Scene Change Detection using Deep Convolutional Neural Networks

Ali Atghaei
*AI Engineer*
*Veunex Company*
*Berlin, Germany*
*Ali.atghaei@veunex.com*

Ehsan Rahnama
*AI Engineer*
*Veunex Company*
*Berlin ,Germany*
*Ehsan@veunex.com*

Kiavash Azimi
*AI Engineer*
*Veunex Company*
*Berlin ,Germany*
*Kiavash@veunex.com*

Hassan Shahbazi
*Data Scientist*
*Veunex Company*
*Berlin, Germany*
*H.shahbazi@veunex.com*

**Abstract**

*Finding and localizing the conceptual changes in two scenes in terms of the presence or removal of objects in two images belonging to the same scene at different times in special care applications is of great significance. This is mainly due to the fact that addition or removal of important objects for some environments can be harmful. As a result, there is a need to design a program that locates these differences using machine vision. The most important challenge of this problem is the change in lighting conditions and the presence of shadows in the scene. Therefore the proposed methods must be resistant to these challenges. In this article, a method based on deep convolutional neural networks using transfer learning is introduced, which is trained with an intelligent data synthesis process. The results of this method are tested and presented on the dataset provided for this purpose. It is shown that the presented method is more efficient than other methods and can be used in a variety of real industrial environments.*

## 1. Introduction

One of the most important issues in the field of intelligent care is the detection of conceptual changes in the scene [1]. For example, when an important object in an environment changes at a certain time, or an object is added or removed from that scene, it must be correctly recognized. For example, in the HSE laws of various industries, the issue of housekeeping is raised in the same context [2]. Or in the case of protecting a place, if an object is added to a scene or removed from it, it must be recognized.

In this problem, one image is considered as the basic image and the other image as the suspicious ones [3,4]. For example, the first image can be the image of an industrial environment at the start of the work shift, and the suspicious image is the image at the end of the work shift.

The goal of the problem is to discover in the second image all the objects that have been changed, deleted or added compared to the base image [4,5]. Because the variety of objects in terms of shape, color, material, texture, and name is very large, and the model must detect their changes in different scenes, it is practically impossible to classify and label objects, and the models in this field only detect changes in the scene. For example, in the application of housekeeping, after identifying and alerting changes, the environmental supervisor checks the items in the regulations of that environment in terms of the presence or absence of certain objects [2]. For instance, a fire extinguisher that should be in a certain place may not have been removed or, a tool must not have been added to the scene, or the passage of people should not be blocked due to the placement of some devices disrupting the start of the next shift. Essentially, the environment at the end of the shift is conceptually similar to the beginning of the shift and the neatness and hygiene of the environment is observed [6].

Perhaps the first solution that comes to mind for this problem is to subtract the two photos, but the change in the lighting conditions of the scenes, the presence of shadows, noises and color changes, undermines this method. There are other similar methods to this problem. Methods such as object recognition or semantic segmentation of images. But due to the previously mentioned challenges related to the variety of objects and complex lighting conditions, the efficiency of these methods is very low in practice [7]. It should also be mentioned that the underlying assumption of this problem is that the camera is fixed in the environment and the shooting angle and the distance and proximity to the scene do not change in the two images (similar to environmental surveillance cameras).

In our previous work on this subject [8], deep neural networks with 6-channel input and 4-channel output were used. But extracting the conceptual feature of more complex structures resistant to the change of different light and visual conditions has determined the need to use deeper pre-learned networks and the use of transfer learning.

Therefore, in this article, a structure is presented for this purpose, and the proposed method is also examined from a theoretical point of view. Also, a procedure for synthesizing training data is proposed and a section is also dedicated to consider the condition of shadowed images with data augmentation.

Two different datasets with different levels of difficulty have been prepared and labeled for testing at the pixel level, and the results of the experiments on these two datasets are presented in the experiments section.

The important innovations of this article include:

- Designing a segmentation model using transfer learning in convolutional neural networks.

- Designing a data augmentation process considering different lighting conditions in the presence of shadows.

## 2. Related Works

In detecting changes in the scene based on two images that were captured at different times from the same place, localizing the changed pixels in terms of the presence or absence of a specific object in that part of the desired image, several previous methods have been investigated.

One of the most challenging factors of this issue is the difference in lighting conditions and the slight changes in shadow conditions in each scenes, and methods should be robust to various changes[9].

In environments that are equipped with a fixed surveillance camera and any change in the movement of objects is captured and the image is crowded the aforementioned changes such as lighting, noise, shadows and camera settings pose important challenges in the problem. A recent important trend in this field is the use of deep convolutional neural networks[10].

Sakura et al. [11] presented a method for localization of changes using image features extracted with CNN by applying contrastive loss. GUO et al. [12] added more resistance to noise and optical changes to CNN features. Lei et al. [13] have presented a deep CNN using image feature association. Khan et al. [14] used deep CNN using Directed Acyclic Graph (DAG) geometrical features of regions to understand conceptual changes. Alcantarilla et al. [15] use a deconvolutional network to detect structural changes in street view images.

Many change detection models have been designed for remote sensing image applications that provide acceptable results on field datasets such as LEVIR-CD and DSIFN-CD [16,17]. One of the mostly used architectures for this problem is to create an encoder-decoder structure. The encoder networks commonly consist of CNN or Transformers [18,19]. In some models, the output of encoder networks are integrated, with some methods such as interaction layer [20], and spatial-temporal interaction [16]. Also the decoder includes layers such as CNN, MLP, and transformer [18,19].

In our previous work [8], this problem has been addressed by connecting two images in the form of one photo with 6 channels (two three-channel photos) and designing a deep CNN network with U-NET structure and providing a data synthesis process. But extracting deep features in this problem requires a large number of of labeled data.

Different methods of transfer learning have greatly helped classification in different problems in the absence of labeled data [21] and unsupervised problems. The use of these methods in segmentation problems such as the SSCD problem can have an important impact.

Methods for abnormal event detection using adversarial deep neural networks have been proposed, which identify the distribution of normal scenes [22], and the basis of these ideas can also be used in the field of detecting the difference in two scenes of this problem.

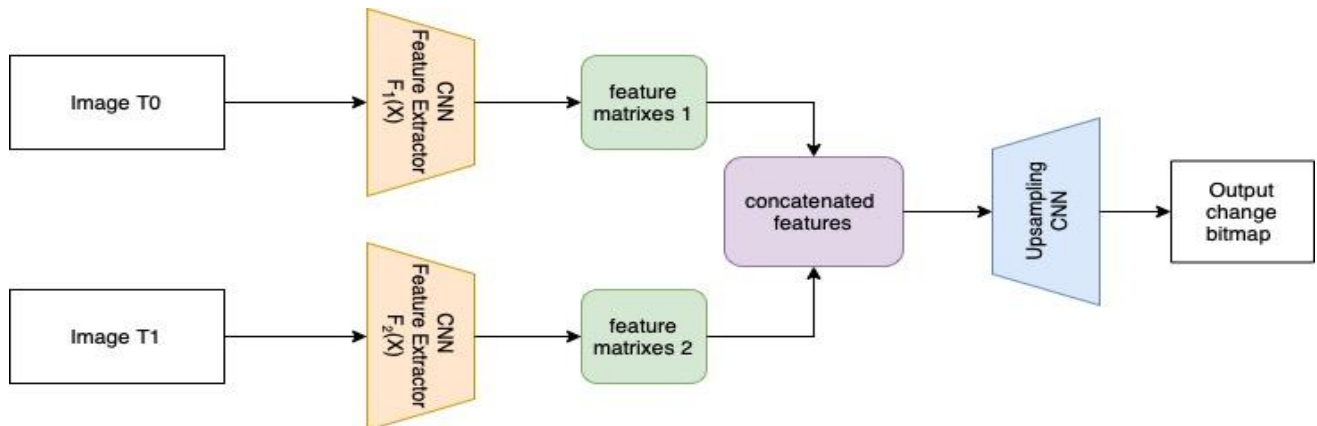

**Figure 1 : Overview of our work. Two feature extractors generate a two feature matrices and the concatenation of those two matrices is fed to the upsampling CNN and the output mask will be constructed.**

The ultimate goal of this problem is segmentation. That is, in the form of a photo, the location of the pixels belonging to the changed objects should be determined. Different methods based on UNET have been widely used in the field of segmentation and have created a suitable accuracy on the dataset of this field [23]. Therefore, it is recommended to use these ideas in the proposed method. Deep Learning algorithms use gradient descent approach to learn the objectives, also known as loss functions [24]. In the field of segmentation, different loss functions have been introduced to increase the accuracy of detection at the object level [25,26]. Our proposed method uses pre-trained deep convolutional networks in a UNET structure and object-wise and pixel-wise loss functions to segment the conceptual changes of two scenes.

## 3. Proposed Method

In this problem, having two images from two different times, the objects that were changed should be identified as the changed area. For this purpose, in our previous work [8], we trained a U-NET based convolutional network, by taking two inputs as a 6-channel tensor (the result of concatenation of the two mentioned images) and the output of a tensor as 4 channels. But in order to evaluate more accurately, new datasets were provided, and by observing the results of previous methods, in order to provide a better method, the proposed method of this article was designed.

The use of transfer learning in various problems reduces the need for a lot of data. Also, the use of pre-trained networks to extract features has been very effective because by seeing a large number of data during training, they extract more important conceptual features [27].

Therefore, according to the definition of this problem, this concept and pre-trained networks can be used. In deep convolutional networks, for example, before the classification layers, the output of the network will be several convolutional maps. Each of these maps consider a specific receptive field on the original image. Therefore, if the features of the first image are extracted with a pre-learned network and some convolutional maps are generated, and then the same is done for the second image of the problem, the convolutional maps can be compared peer-to-peer.

Due to the different light conditions, the feature extraction networks should also be trained according to the problem definition so that the features of the areas of the image that have not changed are extracted in the same way and those that have changed are done differently. For this purpose, according to Figure (1), an upsampling convolutional network is designed, the input of which is the connection of two feature vectors from the first and second image, and its output is one black and white map. These maps were labeled pixel-wise.

Now, in order to train this network, there is a need to provide a rich data set. For this purpose, a labeling process has been defined. The data was provided in the form of connecting two images that were taken from the same place but at different times and light conditions in different environments and in large numbers. To label this data, two images are placed together and the person draws a polyhedron around each of the objects that are in the first image but not in the second (the addition of an object) and vice versa (the removal of an object). After receiving this labeled data, the change label is also obtained using the OR operation of two matrices.

Another process has been used as training data synthesis that is explained in Section 3.1.

### 3.1. Training Data Synthesis

Due to the wide variety of objects present in different industrial places and the impossibility of labeling the names of each of them, the automatic production process is used to prepare the training dataset. In this way, first, a set of pairs of images related to different times of the same scene with different light and shadow conditions are prepared, which are considered unchanged.

In another set, a set of different objects is prepared and their background is removed. Now, according to Figure (2), by adding each of the mentioned objects to one of the pair images, change mask will have a value of 1 in the area of the added object and in the rest of the places it will be labeled 0. By implementing this process, it is possible to prepare any number of training data for use in the problem.

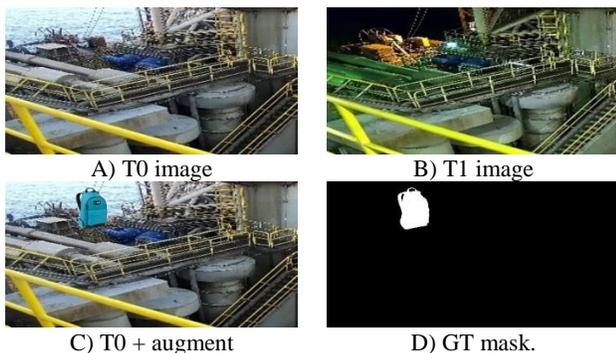

A) T0 image    B) T1 image

C) T0 + augment    D) GT mask.

**Figure 2 : The process of augmenting desired objects and preparing a labeled training dataset. Image A and B are two images of the same scene at different times that have no conceptual difference. In image C, an arbitrary object has been added to image A (the blue bag), resulting in the creation of mask D.**

This method can also be applied to labeled data containing changes in the scene. By adding desired objects to them, the output mask also changes. Adding these desired objects at adjustable rates can be performed randomly anywhere in the scene. Therefore, the number and variety of training data increases.

Another point that should be present in the variety of training data is the presence of shadows. For this purpose, two processes have been considered for data synthesis. First, from the beginning, two scenes from the same environment that differ only in terms of the presence of shadows are selected and the proposed labeling or synthesis process is performed on them. But in the second method as illustrated in Figure (3), it is performed by creating several patterns from the existence of shadows in the form of mask images and randomly selecting a number of them and creating diverse patterns of shadows and adding their weights to the original images while not making any changes to the GT mask. Now, if an area already had a change, with the addition of shadows, it still contains the same change in GT mask, and thus the variety of training data on the subject of the presence of shadows increases.

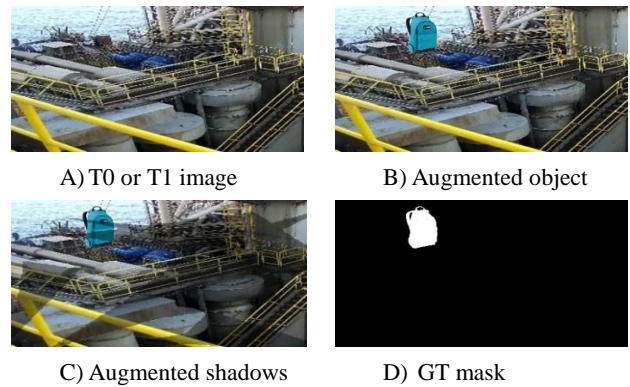

A) T0 or T1 image    B) Augmented object

C) Augmented shadows    D) GT mask

**Figure 3 : The process of shadow augmentation on the image. In step C, a pattern of shadows is randomly selected and added to the image of step B, and has no effect on the GT mask.**

The rest of the typical methods of augmentation, such as noise, resolution change, and optical filters are also used in the problem, which are not explained in detail in this article.

### 3.2 Model Training

According to Figure 1, two images from two different times are entered into two feature extraction deep networks. In designing these networks and initial weights, pre-trained networks such as XceptionNet [28], VGGNet [29], ResNet [30], etc. can be used.

Choosing which layer's output should be used as feature matrix is an important discussion in this issue, which has also been investigated in the experiments section. In general, layers with less depth extract more general features and deeper layers extract more complex features [31]. Therefore, according to the definition of this problem, the appropriate layer should be reached, which is considered in this method as layer 26 of the XceptionNet network. At the end of this step, two sets of feature matrices resulting from applying the network to two images have been created. Every entry of each matrix examines the characteristic of a specific receptive field in two scenes, the corresponding receptive fields in the two scenes usually have high correlation.

As a result, every pixel of each image can affect several elements of feature matrices. Considering that the number of feature matri-ces extracted from each image is large and each one is a description of the meaning of the receptive field of that image, in order to understand the semantic difference of two scenes, the relationship of a large number of elements of the feature matrices of the two images should be checked.

To discover and train these connections, a function is needed, which is implemented using an upsampling deep convolutional network. According to Formula (1) each image enters the feature extractor function and the resulting feature matrices are concatenated together. $g_\theta$ is a function implemented by a neural network where theta represents its parameter set. Upon receiving the combined feature matrices, it converts them into the output mask in which the pixels of the changed area have a value of 1 and the rest of the areas have a value of 0. The labeled and augmented dataset described, is used to train this model.

$$f(I_{T_0}, I_{T_1}) = F_1(I_{T_0}) \oplus F_2(I_{T_1}) \tag{1}$$
$$P_{pred} = g_\theta(f(I_{T_0}, I_{T_1}))$$

The output of this model can be considered as segmentation in which two issues are important next to each other. First, each predicted pixel must be equal to its corresponding GT pixel, and since the value of pixels of GT is limited to zero or one, binary error functions can be used. Second, the detected objects must have large intersection with their corresponding objects in the GT-mask, and each object must be detected separately.

In this regards as stated in Formula (2), the first cost function is considered for pixel accuracy, and the second cost function is considered for different object detection accuracy.

$$L_{seg} = L_{pix} + L_{obj} \tag{2}$$

Recently, in the segmentation articles, the cost functions considering the intersection of the output image with the GT mask showed a favorable performance [32]. One of them is the Dice function [25], which is also selected as L(obj) in this method.

$$L_{dice} = 2 - \frac{2 \times \sum_{i=1}^{N} P_{true_i} \times P_{pred_i}}{\sum_{i=1}^{N} P_{true_i}^2 + \sum_{i=1}^{N} P_{pred_i}^2 + \varepsilon} \tag{3}$$

In the formula (3), $P_{pred}$ means prediction of the model and $P_{true}$ means GT, and N is the number of pixels of the output mask, and the number 2 is chosen for ease of computation. It can be seen that this function checks the cost ratio of intersection to union in the output of the model and GT mask.

Also, in order to consider the pixel accuracy according to whether the value of each pixel is 0 or 1, the binary cross-entropy cost function as illustrated in Formula (4) has been used in this problem.

$$L_{bce} = \frac{1}{N} \sum_{i=1}^{N} -(P_{true_i} \times \log(P_{pred_i}) + (1 - P_{true_i}) \times \log(1 - P_{pred_i})) \tag{4}$$

As a result, the overall segmentation cost function is obtained with the Formula (5).

$$L_{seg} = \frac{1}{N} \sum_{i=1}^{N} -(P_{true_i} \times \log(P_{pred_i}) + (1 - P_{true_i}) \times \log(1 - P_{pred_i})) \tag{5}$$
$$+ (2 - \frac{2 \times \sum_{i=1}^{N} P_{true_i} \times P_{pred_i}}{\sum_{i=1}^{N} P_{true_i}^2 + \sum_{i=1}^{N} P_{pred_i}^2 + \varepsilon})$$

This way, the convolutional networks used in this method are trained with the gradient of these cost functions.

In the inference process, by limiting the contours of connected components based on their area, it is possible to avoid considering small objects in the output.

## 4. Experiments

As we explained in the proposed method the challenge is finding the difference between two scenes. We evaluate our method in this section on our dataset, including real world image-pairs. Sections 4.1 to 4.3 explain the experimental setup and Sections 4.4 to 4.8 describe the ablation study and comparison the proposed method with other existing methods in this research field.

### 4.1 Dataset Setting

To prepare the dataset, we designed a game similar to a puzzle in such a way that we used paired images of the same scene at a different time and drew polygons wherever there were regions of semantic difference pixels. Due to the lack of data in industrial environments and the difficulty of collecting it, we solved this deficiency by synthesizing data on our limited gathered dataset in the training phase. Therefore, we provided 9404 image pairs in different light conditions from indoor and outdoor industrial environments. For training, we used 1096 real world image pairs and 8000 synthetic image pairs. For the testing phase, 308 image pairs just in real world domain were chosen. We demonstrated the distribution of the dataset in Table 1.

| Dataset | Domain | Number of pairs | Annotation |
|---|---|---|---|
| Train | Real + Synthesis | 1096 + 8000 | Pixel level |
| Test without shadow | Real | 156 | Pixel level |
| Test with shadow | Real | 152 | Pixel level |

Table 1. **Train and test dataset description.**

### 4.2 Implementation Details

For a practical deep learning system, "The devil is always in the details". Our implementation is based on the general framework Tensorflow [32]. We used pre-trained Xception [28] weight for the features extractor and designed upsample parts by CNN layers. During training, Xception layers were fine-tuned, and computing the gradient for upsample layers. Adam optimizer [33] is used for training with a learning rate of 0.001. We employ the

"poly" learning rate policy [34] to schedule learning rate. The initial learning rate is multiplied by $(1-\frac{iter}{max\_iter})^{0.9}$.

The batch size was 8, and the image size was 256 × 512 pixels. For data augmentation, we adopt change of contrast and brightness by the severity of 0.3 and median blur with kernel size 3 for all datasets, Additionally shadow augmentation was also implemented. The network is trained on two 1080TI GPUs.

### 4.3 Evaluation Metrics

According to previous studies [35] pixel-level metrics, such as mean pixel accuracy or mean Intersection over Union have been widely used. However, these metrics are not suitable to evaluate object-level scene change detection. For pixel-level metrics, errors in small objects have little impact and they are often neglected compared to larger objects [36]. Therefore, we employed the Precision, Recall, and F1 score, which is widely used in object detection tasks, all of which have a maximum value of 1. In high risk industrial places, such as locations with explosives on the scene, alerting the HSE supervisor is crucial. In this case, the recall value is a better metric to indicate the performance of the model. In contrast in places where the risk is less, but the environment must be neat, precision is a better criterion. Generally, F1-score is the balance of precision and recall because it can be interpreted as a harmonic mean of the two. Therefore, it is a better criterion for the assessment of the model. We calculated the metrics as follows. We compared the prediction mask with the ground truth (GT) and determined the accordingly true positive (TP) or false positive (FP). Any GT regions that the model failed to detect were classified as false negatives (FN).

$$Precision = \frac{TP}{TP+FP}$$

$$Recall = \frac{TP}{TP+FN}$$

$$F_1 = \frac{2 \times Precision \times Recall}{Precision + Recall}$$

### 4.4 Different Pre-trained Models as Backbone

In this section, we compared different pre-trained models that were trained on ImageNet as feature extractors. In this study, we used VGG19 [29], ResNet50 [30], and Xception [28] networks with different layers as the backbone of our network Finally, we present the results of this survey on our dataset in Table 2. As shown in Table 2 the result indicated the pre-trained Xception network with the 26[th] layer as feature extractor is proper for our case. As seen in the table, when pre-trained models are compared to themselves, the model with low depth has high precision and recall except ResNet50. The high values of precision and recall for ResNet50 are related to the 52[nd] and 17[th] layers respectively. As Stated in Section 4.3, the F1-Score is a suitable metric, therefore Xception with the 26[th] layer as FE has the highest value among all.

In addition, we evaluated our network by removing such conditions that were reported in Section 4.5 Ablation study, and comparison with the previous state-of-the-art network are presented in Section 4.7.

| Pre-trained Model | FE layer | Precision | Recall | F1 Score |
|---|---|---|---|---|
| VGG19 | 11 | 0.75 | 0.50 | 0.60 |
| VGG19 | 17 | 0.63 | 0.44 | 0.51 |
| ResNet50 | 17 | 0.47 | 0.75 | 0.54 |
| ResNet50 | 52 | 0.83 | 0.45 | 0.58 |
| **Xception** | **26** | **0.72** | **0.61** | **0.66** |
| Xception | 66 | 0.74 | 0.51 | 0.60 |

Table 2. **Comparison of different feature extractors impact on evaluation metrics**

### 4.5 Ablation Study about Shadow Augmentation

We use the model in industrial environments at different hours of the day and night. For this reason, we have shadows on objects during the day and it probably changes the pattern of objects. It is possible that the model considers it as changes even though no changes have been made in the scene. To evaluate the performance of the model in these fluctuated conditions we remove shadow augmentation and different lighting conditions from the training dataset on the first try and add them to the dataset on the next try.

| Shadow Augmentation | Precision | Recall | F1-Score |
|---|---|---|---|
| No Shadow | 0.68 | 0.47 | 0.55 |
| Shadow | 0.61 | 0.67 | **0.61** |

Table 3. **Ablation study on shadow augmentation**

As seen in Table 3, when we remove shadow augmentation from the dataset compared to the state we have, recall decreases by 29.8%. The slight decrease in precision, can be neglected because of the high F1-score value. The performance was improved when data was mixed with shadow augmentation and the F1-score increased by 15%.

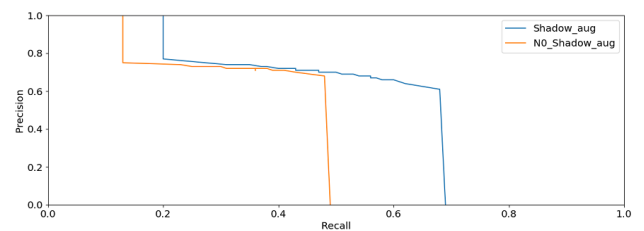

Figure 4. **Comparison of different types of data augmentation illustrates the effectiveness of shadow augmentation in this problem.**

Figure 4 demonstrats a precision-recall curve. As is clear, the area under the shadow augmentation curve is more than the case with no shadow augmentation and the recall is significantly higher.

## 4.6 Ablation Study about Xception Trainable Layers

We evaluate different network settings. In this regards, we used different layers of Xception to be trainable, and reported the impact of this in Table 4. We did this because the Xception model was trained with the ImageNet dataset and the objects we have in the industrial scene could have a different pattern. Therefore, we compute gradients for all layers, the last five layers, and the last three layers of the Xception model.

The precision of the last three layers is barely better than the last five and all layers. In Table 4, we reported precision and recall values in the highest F1-score value. As seen in the precision-recall curve Figure 5, the area under the curve (AUC) for the last three layers is more than others. Therefore it demonstrates the performance of the model would be better in this state. As shown in Table 4, The recall value for all tests is 0.69 but the precision value for the last three layers is 0.68 and is roughly 0.01 is more than others.

| Trainable Layers | Precision | Recall | F1-Score |
|---|---|---|---|
| All layers | 0.67 | 0.69 | 0.67 |
| **Last 3 layers** | **0.68** | **0.69** | **0.68** |
| Last 5 layers | 0.67 | 0.69 | 0.67 |

Table 4. **Comparison of trainable layers of the Xception model on our dataset.**

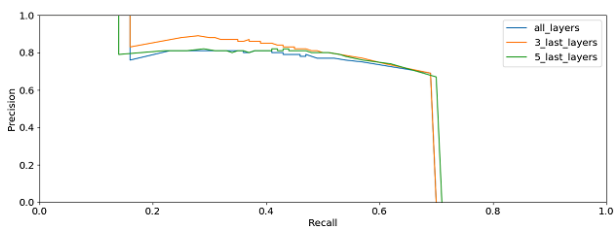

Figure 5. **Precision-recall curve comparison about different trainable layers of XceptionNet in this method as feature extractor.**

## 4.7 Ablation Study about Loss Functions

As we explained in Section 3.2, for the loss function we combined binary cross entropy (BCE) and Dice for pixels and objects loss respectively. In this section, we inquire about the impact of removing each of these loss functions on the performance of the model.

As seen in Figure 6 and Table 5, while the loss function is BCE recall is 0.88 standing better than others, precision is very low. It shows there are high false positive. When only Dice is used in loss function, the precision value increases by 25.9% but PR-curve shows that the AUC of dice is lower than BCE so it does not perform well. But when we use the combination of these two as a cost function, we obtain better performance. The precision value is 0.72 and the recall is 0.69. The precision has increased showing that the false positive is decreased. In this state, the F1-score remarkably changed. It increases by 21.4% and 17.1% compared to BCE and Dice respectively.

| Loss Function | Precision | Recall | F1-Score |
|---|---|---|---|
| Binary-Cross-Entropy (BCE) | 0.40 | 0.88 | 0.55 |
| Dice | 0.54 | 0.64 | 0.58 |
| BCE + Dice | **0.72** | **0.69** | **0.70** |

Table 5. **Effect of combination of Dice and BCE loss function on results of this method.**

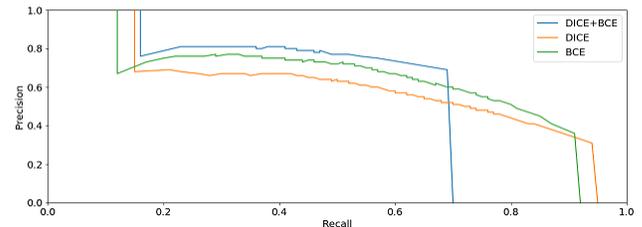

Figure 6. **Precision-recall curves of different loss function. It can be infered that the combination of two loses has a larger AUC and better performance in this problem.**

## 4.8 Comparison to Other Methods

In this section, we compared our method with previous works in this field. We use ChangeFormer [18], and Changer [20] networks for this comparison. These methods are based on deep learning as well. ChangeFormer network uses transform-based architecture and also Changer on the other hand, uses feature interaction layers in this network to find correlation before future fusion. Finally, Table 6 shows the results of our method compared with others on our dataset. The F1-score of our method has increased by 50% and 15% compared to Changer and ChangeFormer networks respectively on our dataset. Also the precision of our method is slightly higher than others. The precision shows that we have fewer false positive than others.

| Methods | Precision | Recall | F1 Scre |
|---|---|---|---|
| Changer | 0.68 | 0.33 | 0.44 |
| ChangeFormer | 0.62 | 0.53 | 0.57 |
| Ours | **0.72** | **0.61** | **0.66** |

Table 6. **Comparison of our method with other previous methods on our industrial dataset.**

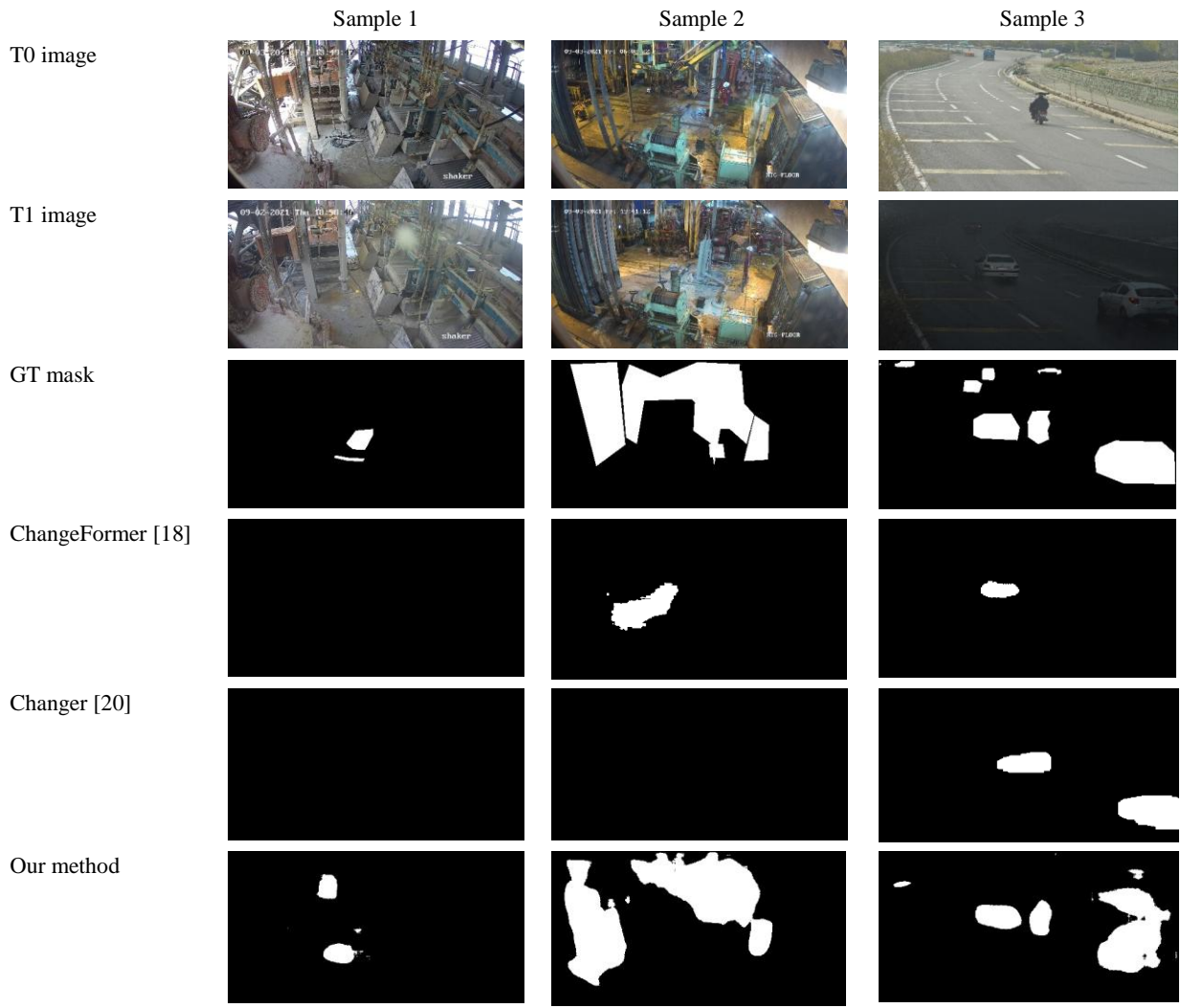

Figure 7. **Comparing the results of our model and two of the important models in this field on three samples of image pairs shows that the output mask of our model was more similar to the GT mask.**

As shown in Figure 7, in crowded industrial images and in conditions where light changes and the presence of shadows are significant, our model has closer results to the GT mask. As stated in Section 1, the model's resistance to light changes is very important in industrial applications, so our model has more usability than other models in this field.

## 5. Conclusion

In this paper, we proposed a novel framework for scene change detection. The main idea is to learn a feature representation to find and localize the change area. This method segments the differences of two images belong to same scenes but in different time with illumination change and different shadow conditions. To this end we also propose a novel augmentation proce dure to help the model being robust on this illumination changes conditions. Extensive experimental results on industrial scene change detection, with large variety of objects and shadow conditions, demonstrate that our proposed method is able to find changes with better performance in comparison with other methods in this field.